\documentclass[letterpaper]{article} % DO NOT CHANGE THIS
\usepackage{aaai23}  % DO NOT CHANGE THIS
\usepackage{times}  % DO NOT CHANGE THIS
\usepackage{helvet}  % DO NOT CHANGE THIS
\usepackage{courier}  % DO NOT CHANGE THIS
\usepackage[hyphens]{url}  % DO NOT CHANGE THIS
\usepackage{graphicx} % DO NOT CHANGE THIS
\usepackage{natbib}  % DO NOT CHANGE THIS AND DO NOT ADD ANY OPTIONS TO IT
\usepackage{caption} % DO NOT CHANGE THIS AND DO NOT ADD ANY OPTIONS TO IT
\usepackage{algorithm}
\usepackage{algorithmic}

\usepackage{newfloat}
\usepackage{listings}
\usepackage{newfloat}
\usepackage{listings}
\usepackage{bbding}
\usepackage{times}
\usepackage{amsmath}
\usepackage{amssymb}
\usepackage{enumerate}
\usepackage{mathrsfs}
\usepackage{array}
\usepackage{multirow}
\usepackage{booktabs}
\usepackage{microtype}      % microtypography
\usepackage{makecell}
\DeclareCaptionStyle{ruled}{labelfont=normalfont,labelsep=colon,strut=off} % DO NOT CHANGE THIS
\floatstyle{ruled}
\newfloat{listing}{tb}{lst}{}
\floatname{listing}{Listing}
\title{Unsupervised Hierarchical Domain Adaptation for Adverse Weather Optical Flow}
\author{
		Hanyu Zhou\textsuperscript{\rm 1},
		Yi Chang\textsuperscript{\rm 1}\thanks{Corresponding Author},
		Gang Chen\textsuperscript{\rm 2},
		Luxin Yan\textsuperscript{\rm 1},
}
\affiliations{
		\textsuperscript{\rm 1} National Key Laboratory of Science and Technology on Multispectral Information Processing,
		School of Artificial Intelligence and Automation, Huazhong University of Science and Technology\\
		\textsuperscript{\rm 2} School of Computer Science and Engineering, Sun Yat-sen University\\
		hyzhou@hust.edu.cn, yichang@hust.edu.cn, cheng83@mail.sysu.edu.cn, yanluxin@hust.edu.cn
}

\usepackage{bibentry}
\begin{document}
\maketitle

\begin{abstract}
Optical flow estimation has made great progress, but usually suffers from degradation under adverse weather. Although semi/full-supervised methods have made good attempts, the domain shift between the synthetic and real adverse weather images would deteriorate their performance. To alleviate this issue, our start point is to unsupervisedly transfer the knowledge from source clean domain to target degraded domain. Our key insight is that adverse weather does not change the intrinsic optical flow of the scene, but causes a significant difference for the warp error between clean and degraded images. In this work, we propose the first unsupervised framework for adverse weather optical flow via hierarchical motion-boundary adaptation. Specifically, we first employ image translation to construct the transformation relationship between clean and degraded domains. In motion adaptation, we utilize the flow consistency knowledge to align the cross-domain optical flows into a motion-invariance common space, where the optical flow from clean weather is used as the guidance-knowledge to obtain a preliminary optical flow for adverse weather. Furthermore, we leverage the warp error inconsistency which measures the motion misalignment of the boundary between the clean and degraded domains, and propose a joint intra- and inter-scene boundary contrastive adaptation to refine the motion boundary. The hierarchical motion and boundary adaptation jointly promotes optical flow in a unified framework. Extensive quantitative and qualitative experiments have been performed to verify the superiority of the proposed method.
\end{abstract}

\section{Introduction}

\begin{figure}[t]
	\centering
	\includegraphics[width=0.99\linewidth]{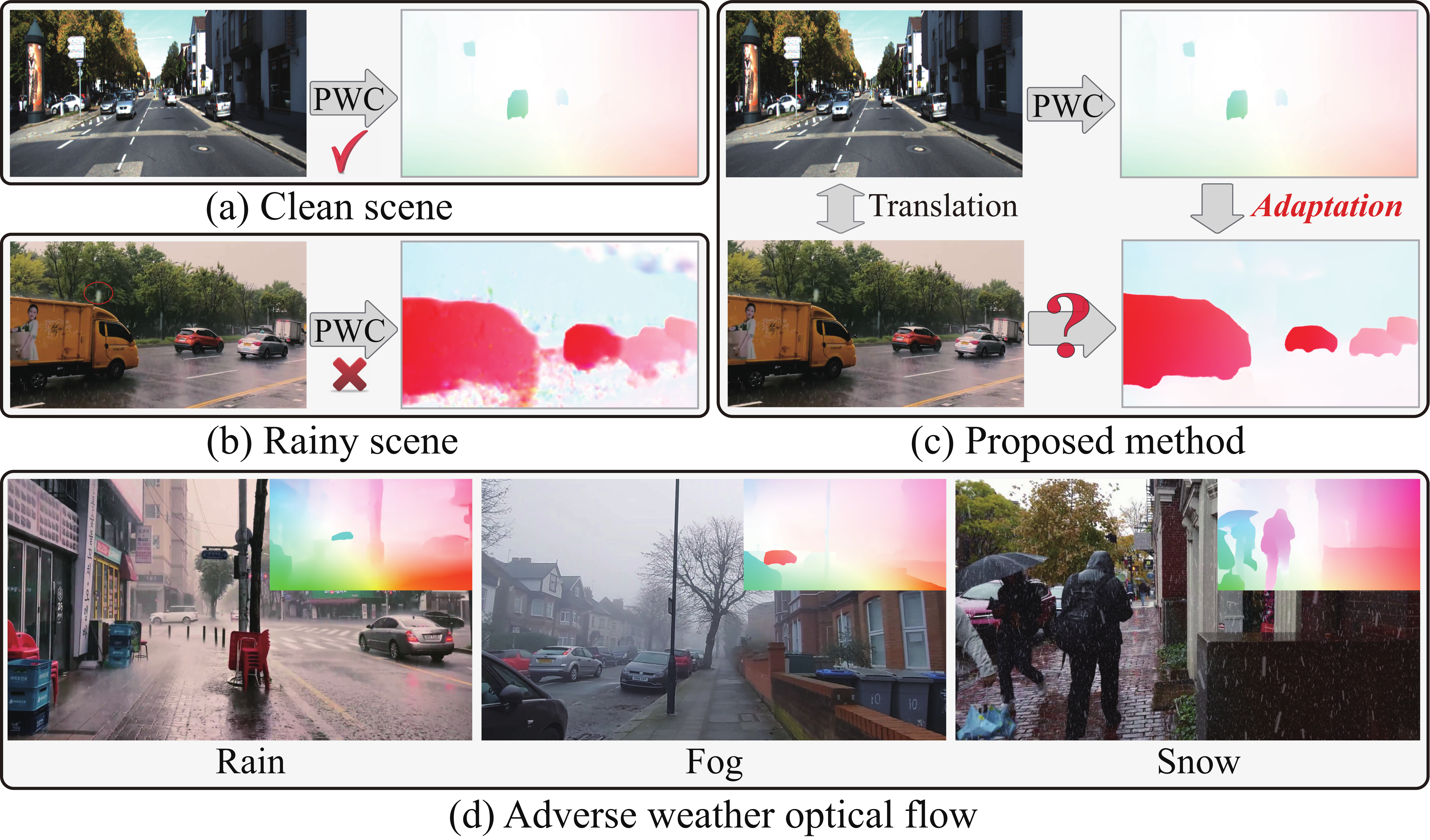}
	\caption{(a) and (b) Classical PWC-Net obtains good flow on the clean scene yet fails to produce satisfactory optical flow under the real degraded scene. (c) The proposed method HMBA-FlowNet bypasses the difficulty of directly estimating adverse weather optical flow, and proposes the first unsupervised domain adaptation method to transfer the knowledge from clean domain. (d) Optical flow could be estimated by the proposed method under various real adverse scenes.}
	\label{RealResults}
\end{figure}

Optical flow is an important task in computer vision, which aims to estimate per-pixel motion in a video sequence and has been widely applied for autonomous driving \cite{sun2020scalability}, object tracking \cite{fan2019lasot}. Existing optical flow methods have achieved great progress under good weather conditions in Fig. \ref{RealResults} (a), but may suffer from challenges under adverse weather in Fig. \ref{RealResults} (b), such as rain and fog. The main reason is that degradation factors have unexpectedly violated the brightness and gradient constancy assumptions. The goal of this work is to estimate optical flow under adverse weather, such as rain, fog, and snow as shown in Fig. \ref{RealResults} (d).

The existing methods have made attempts to get rid of the negative influence of adverse weather for optical flow estimation. The optimization-based optical estimation method \cite{li2018robust} focuses on specific weather conditions rainy or foggy with sophisticated hand-crafted prior, which makes them less suitable for various adverse weather. The learning-based methods require the ground truth labels of clean image, optical flow \cite{li2019rainflow} or auxiliary gyroscope information \cite{li2021gyroflow}, which are not easy to be acquired in real-world scenes. Although previous learning-based methods have achieved great progress, the domain shift between the synthetic and real adverse weather images restricts their performance in real adverse scenes.

To overcome the domain shift issue, our start point is to formulate this task as unsupervised domain adaptation where we can transfer motion knowledge from source clean domain to target degraded domain in Fig. \ref{RealResults} (c). To our knowledge, we are the first to handle adverse weather optical flow in an unsupervised framework. \emph{Our key insight is that the inherent motion is invariant but the warp error is variant between clean and degraded images}. This intuitive observation is reasonable.
On one hand, the optical flow of the same scene should be essentially consistent no matter what weather conditions are. On the other hand, the warp error measuring the alignment of the image structure of two adjacent frames is very small for the clean image while the warp error for the degraded image would be enormous due to degradation. The motion consistency and boundary inconsistency between both domains provide us the powerful knowledge for adaptation.

In this work, we first formulate the degraded optical flow estimation into an ill-posed problem via \emph{maximum-a-posterior}. The mathematical formulation offers us a heuristic guidance on how to principally design the network and unsupervised loss. Consequently, we propose an unsupervised hierarchical motion-boundary adaptation (HMBA-FlowNet) method for optical flow under adverse weather, which consists of three main parts: clean and degraded image translation, motion adaption, and boundary adaptation.

Specifically, we first learn the unpaired image translation between the clean and degraded domains. In the motion adaptation step, we encode the cross-domain clean-degraded images into a motion-invariance common feature space, and then decoder them into the same optical flow via the consistency loss. The motion adaptation could capture the cross-domain flow consistency relationship and transfer the motion information from the clean domain to the degraded domain, thus encouraging preliminary optical flow results for the real degraded images. In the boundary adaptation step, we propose a joint intra- and inter-scene boundary contrastive adaptation in the warp error space. The intra-scene boundary adaptation is to minimize the discrepancy between different patches in the same scene and maximize the distance between the clean and degraded domains, meanwhile, the inter-scene boundary contrastive adaptation encourages the feature manifolds of different scenes in the same domain to be closer to each other. Both intra- and inter-scene boundary contrastive adaptation boost the model to effectively distinguish the boundary from the warp error caused by degradation, thus refining the inherent motion boundary in the real degraded domain. We summarize the main contributions as follows:

\begin{itemize}
\item We design a joint motion-consistency and boundary-inconsistency alignment mechanism for the adaptation, and propose an unsupervised hierarchical motion-boundary adaptation framework for adverse weather optical flow. This is a pioneer exploration to transfer the reliable knowledge from clean domain to adverse weather domain from the unsupervised domain adaptation paradigm.
\item We propose a motion consistency adaptation (MCA) to learn the cross-domain motion-invariance feature which acts as a fidelity to ensure the preliminary motion, and a boundary contrastive adaptation (BCA) to effectively distinguish the motion boundary from the warp error caused by degradation which works as a regularization to refine the inherent motion boundary. The joint adaptation mechanism benefits obtaining better optical flow.
 \item The proposed method consistently outperforms the competing methods on real datasets by a large margin. Moreover, the proposed method generalizes well for different adverse weather conditions, such as rain, fog, and snow.
\end{itemize}

\section{Related Work}
\noindent
\textbf{Optical Flow under Adverse Weather.} In recent years, the learning-based optical flow approaches \cite{dosovitskiy2015flownet, ren2017unsupervised, sun2018pwc, luo2021upflow, goodfellow2014generative, liu2019ddflow} have made great progress.
Although these methods have achieved satisfactory results in clean scenes, they would suffer from degradation under adverse weather. An intuitive solution to this challenging task is to perform the image deraining \cite{fu2017clearing, yang2017deep, zhang2018density} or dehazing \cite{ren2018gated, liu2019griddehazenet} with subsequent optical flow estimation. However, existing restoration methods are not designed for optical flow and possible over-smoothness or residual artifacts may result in a negative influence.

RobustFlow \cite{li2018robust} was the first optical flow estimation method in rainy scenes by constructing the handcrafted residue channel and its colored version prior that is invariant to rain streaks. Further, to improve feature representation, Li \emph{et al} \cite{li2019rainflow} proposed to automatically learn the rain- and veiling-invariant features in a supervised manner. Given that it is difficult to collect ground truth for real scenes, Yan \emph{et al}. \cite{yan2020optical} presented a semi-supervised method DenseFogFlow, and its key idea is to model the flow consistency between clean images and corresponding rendered foggy images. GyroFlow \cite{li2021gyroflow} resorted to a hardware scheme to handle adverse weather, and innovatively utilized the gyroscope to obtain ego-motion labels of the camera for weakly-supervised optical flow estimation. Compared with previous methods, the proposed method is the first to handle the optical flow under adverse weather from an unsupervised perspective which generalizes well for various real adverse weather images.

\noindent
\textbf{Domain Adaptation.} Domain adaptation \cite{chen2021semi} aims to effectively tackle the distribution discrepancy between the source and target domains, which has been widely applied in classification \cite{chen2021hsva}, detection \cite{xie2021detco} and semantic \cite{gao2022cross, ma2022both}. In fact, the previous methods have utilized different invariance properties \cite{li2019rainflow, yan2020optical} between the clean and specific degraded images, which aim to transfer motion knowledge from the source clean domain to the target adverse weather domain. Compared with previous methods, the proposed method not only explores the motion-consistency for preliminary motion estimation, but also excavates the boundary-inconsistency prior to better refine the motion boundary. Our hierarchical motion-boundary adaptation empowers us with better knowledge transfer ability in a unified coarse-to-fine optical flow framework.

\begin{figure*}[t]
	\setlength{\abovecaptionskip}{8pt}
   \setlength{\belowcaptionskip}{-3pt}
	\centering
	\includegraphics[width=0.98\linewidth]{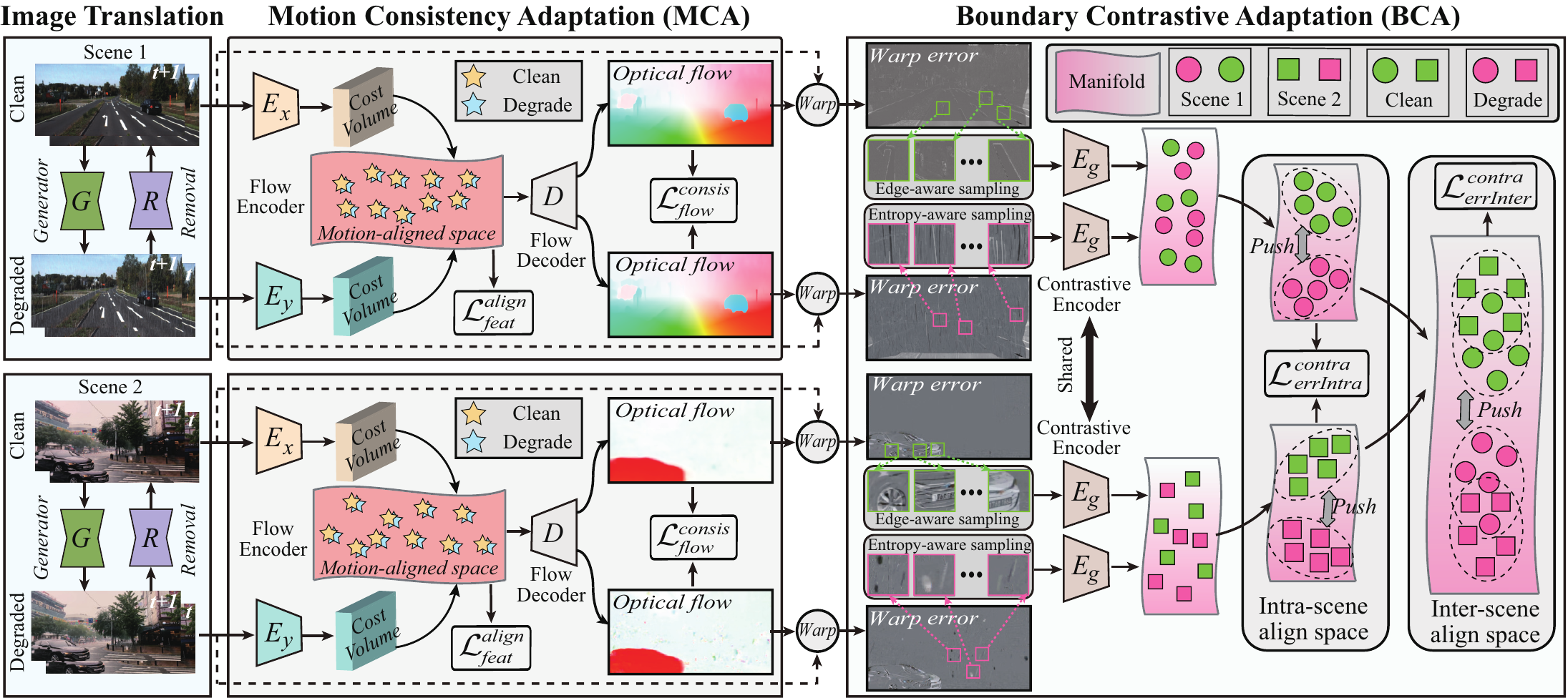}
	\caption{The overall architecture mainly contains image translation, motion-boundary adaptation. We first employ the image translation between clean and degraded domains. In motion adaptation, we align the cross-domain motion features into a common space, and then enforce the optical flow consistency loss to transfer motion knowledge. In boundary adaptation, we propose intra- and inter-scene boundary contrastive adaptation to model the warp error inconsistency to further improve motion boundary.}
	\label{Framework}
\end{figure*}

\section{Hierarchical Adaptation for Optical Flow}
\subsection{Problem and Motivation}
The goal of this work is to estimate optical flow under real adverse weather scenes. Since adverse weather degradation breaks the brightness constancy and gradient constancy assumptions of optical flow, most of the existing optical flow methods usually suffer from degradation under real adverse weather scenes.
To alleviate this issue, we start with the mathematical formulation for the degraded optical flow to reveal the optical flow consistency and warp error inconsistency priors. And then, we propose an unsupervised hierarchical motion-boundary domain adaptation for optical flow under adverse weather scenes in Fig. \ref{Framework}. We first employ the unpaired image translation between the clean domain and the degraded domain. Consequently, we transfer motion knowledge from the clean domain to the degraded domain via motion adaptation, and align the distribution of motion boundary of intra- and inter-scene via boundary adaptation.

\subsection{Mathematical Formulation}
Given an adverse weather image \textbf{\emph{Y}}, we simply assume the degradation as the linear additive decomposition model:
\begin{equation}
\footnotesize
  \setlength\abovedisplayskip{2pt}
  \setlength\belowdisplayskip{2pt}
  \textbf{\emph{Y}} = \textbf{\emph{X}} + \textbf{\emph{D}},
  \label{eq:linear_model}
\end{equation}
where \textbf{\emph{X}}, \textbf{\emph{Y}}, and \textbf{\emph{D}} denote clean image, degraded image, and adverse component, respectively. The motion is usually defined as the linear additive relationship of both the foreground object and background scene \cite{lv2018learning}. We assume that clean image \textbf{\emph{X}} and adverse component \textbf{\emph{D}} are independent from each other. For example, the motion of the rain/snow is independent from the foreground object and background. Thus, we enforce optical flow operator \emph{F} on Eq. (\ref{eq:linear_model}):
\begin{equation}\footnotesize
  \setlength\abovedisplayskip{2pt}
  \setlength\belowdisplayskip{2pt}
  \emph{F}(\textbf{\emph{Y}}) = \emph{F}(\textbf{\emph{X}}) + \emph{F}(\textbf{\emph{D}}),
  \label{eq:opticalflow_model}
\end{equation}
That is to say, the optical flow of the degraded image can be inferred from that of the clean image. Thus, we can formulate the optical flow under adverse weather as an ill-posed inverse problem via \emph{maximum-a-posterior} \cite{murphy2012machine}:
\begin{equation}\scriptsize
	\setlength\abovedisplayskip{2pt}
  \setlength\belowdisplayskip{2pt}
\mathop {\min }\limits_{X,R,F} \underbrace {\gamma ||F(\textbf{\emph{Y}}) - F(\textbf{\emph{X}})|{|_1}}_{\footnotesize{flow{\kern 1pt} {\kern 1pt} consistent}} + \underbrace {\delta {P_x}(F(\textbf{\emph{X}}))}_{\footnotesize{clean{\kern 1pt} {\kern 1pt} flow}} + \underbrace {\rho {P_d}(F(\textbf{\emph{D}}))}_{\footnotesize{degraded{\kern 1pt} flow}} {\kern 1pt}{\kern 1pt}{\kern 1pt} \emph{s.t.} {\kern 1pt}{\kern 1pt}{\kern 1pt} \textbf{\emph{X}} = R(\textbf{\emph{Y}}),
  \label{eq:Constrained MAP_model}
\end{equation}
where \emph{R} is the image translation, the first term is flow consistency between clean and degraded images, the second and third terms ${P_x}$, ${P_d}$ denote the optical flow prior knowledge of clean image and degradation, respectively. We further relax the constrained problem Eq. (\ref{eq:Constrained MAP_model}) into an unconstrained one:
\begin{equation}\footnotesize
	\setlength\abovedisplayskip{2pt}
  \setlength\belowdisplayskip{2pt}
	\begin{aligned}
\mathop {\min }\limits_{X,R,F} \underbrace {\alpha ||\textbf{\emph{X}} - R(\textbf{\emph{Y}})|{|_1}}_{\scriptsize{image{\kern 1pt} {\kern 1pt} translate}} &+ \underbrace {\beta {P}(\textbf{\emph{X}})}_{\scriptsize{image}}  +   \underbrace {\gamma ||F(\textbf{\emph{Y}}) - F(\textbf{\emph{X}})|{|_1}}_{\scriptsize{flow{\kern 1pt} {\kern 1pt} consistent}} \\
&+ \underbrace {\delta {P_x}(F(\textbf{\emph{X}}))}_{\scriptsize{clean{\kern 1pt} {\kern 1pt} flow}} + \underbrace {\rho {P_d}(F(\textbf{\emph{D}})),}_{\scriptsize{degraded{\kern 1pt} flow}}
  \label{eq:Unconstrained MAP_model}
\end{aligned}
\end{equation}
where the first two terms are to translate the image between clean and degraded domains, and the intention of the third and fourth terms is to estimate the optical flow from clean domain and construct motion adaptation between both the domains, and the fifth term is the flow prior about the degradation. $\alpha, \beta, \gamma, \delta, \rho$ are balance hyper-parameters. Next, we will describe how we design each term and overall network to mimic the optimization function Eq. (\ref{eq:Unconstrained MAP_model}).

\subsection{Unpaired Image Translation}
\label{sec:image_translation}
Our key idea is to transfer motion knowledge from clean to degraded domain. Thus, the image translation between clean and degraded domains is essential to build the relationship between them. Intuitively, we can employ a classical generative network \cite{isola2017image} to mimic the first two terms in Eq. (\ref{eq:Unconstrained MAP_model}). In this work, instead of using the conventional GAN, we employ the unpaired CycleGAN \cite{zhu2017unpaired}, which takes two unpaired scenes as input for robust image translation and abundant knowledge transfer.
Thus, given two adjacent clean images $\textbf{\emph{X}} = {[\textbf{\emph{X}}_1, \textbf{\emph{X}}_2]}$ and degraded images $\textbf{\emph{Y}} = {[\textbf{\emph{Y}}_1, \textbf{\emph{Y}}_2]}$, the image translation loss can be expressed as:
\begin{equation}
	\setlength\abovedisplayskip{3pt}
  \setlength\belowdisplayskip{3pt}
\mathcal{L}^{tran}_{img} = ||R(G(\textbf{\emph{X}})) - \textbf{\emph{X}}||_1 +  ||G(R(\textbf{\emph{Y}})) - \textbf{\emph{Y}}||_1,
 \label{eq:imageconsistency}
\end{equation}
where \emph{R}, \emph{G} is the degraded-to-clean and clean-to-degraded image translation networks, respectively. We follow \cite{zhu2017unpaired} to express the image prior with adversarial loss:
\begin{equation}
	\setlength\abovedisplayskip{3pt}
  \setlength\belowdisplayskip{3pt}
\begin{aligned}
\mathcal{L}^{prior}_{img} = \mathbb{E}_{x}[{logD_{x}(\textbf{\emph{X}})}] +  \mathbb{E}_{y}[{log(1-D_{x}(R(\textbf{\emph{Y}})))}] \\
+ \mathbb{E}_{y}[{logD_{y}(\textbf{\emph{Y}})}] +  \mathbb{E}_{x}[{log(1-D_{y}(G(\textbf{\emph{X}})))}],
 \label{eq:Adversarial}
 \end{aligned}
\end{equation}
where $D_x$ and $D_y$ are discriminators to make the images lying on the natural clean and degraded manifolds, respectively.

\subsection{Motion Consistency Adaptation}
\label{sec:Optical}
According to Eq. (\ref{eq:Unconstrained MAP_model}), the third and fourth terms enlighten us to bypass the difficulty to solve this challenging problem in the degraded domain, but instead, we can resort to the motion knowledge in the clean domain. The physical meaning of the two terms is to transfer optical flow from clean domain to degraded domain via the motion consistency. We employ the simple yet effective PWC-Net \cite{sun2018pwc} as our backbone to learn optical flow. We first take two encoders $E_x$ and $E_y$ after the cost volume layer to align the cross-domain inputs into a motion-invariance common feature space. It is worth noting that, although we share the same architecture for optical flow of clean and degraded domains, their weights of encoders have better not be shared (please refer to the discussion). We enforce the cost volume feature consistency loss to align the correlation features between both domains:
\begin{equation}
	\setlength\abovedisplayskip{3pt}
  \setlength\belowdisplayskip{3pt}
\begin{aligned}
\mathcal{L}^{align}_{feat} = ||{C(E_y(G(\textbf{\emph{X}}))) - C(E_x(\textbf{\emph{X}}))}||_1 \\+ ||{C(E_y(\textbf{\emph{Y}})) - C(E_x(R(\textbf{\emph{Y}})))}||_1
  \label{eqa:correlation_consistency_loss},
\end{aligned}
\end{equation}
where $C(\cdot) $ is the cost volume layer that extracts the most discriminative motion features. The motion-aligned embedding features could provide a homogeneous adaptation preparation for the image-level flow adaptation. Consequently, we enforce one shared flow decoder $D$ to estimate optical flows of both domains from the motion-aligned feature space with the optical flow consistency loss as follows:
\begin{equation}
\setlength{\abovedisplayskip}{3pt}
\setlength{\belowdisplayskip}{3pt}
\begin{aligned}
\mathcal{L}^{consis}_{flow} = ||{D(C(E_y(G(\textbf{\emph{X}})))) - D(C(E_x(\textbf{\emph{X}})))}||_1 \\
+ ||{D(C(E_x(R(\textbf{\emph{Y}})))) - D(C(E_y(\textbf{\emph{Y}})))}||_1.
  \label{eqa:flow_consistency_loss}
\end{aligned}
\end{equation}
Eq. (\ref{eqa:correlation_consistency_loss}) and Eq. (\ref{eqa:flow_consistency_loss})  jointly achieve the flow consistency adaptation of the third term in Eq. (\ref{eq:Unconstrained MAP_model}). As for the clean optical flow prior, namely the fourth term in Eq. (\ref{eq:Unconstrained MAP_model}), the brightness and gradient constancy assumptions still hold for clean images. Thus, we employ the photometric loss \cite{jason2016back} to compute the brightness difference between two adjacent images $\textbf{\emph{X}}_1$, $\textbf{\emph{X}}_2$ for non-occluded pixels:
\begin{equation}
	\setlength{\abovedisplayskip}{3pt}
	\setlength{\belowdisplayskip}{3pt}
\begin{aligned}
\resizebox{0.1\hsize}{!}{$\mathcal{L}^{prior}_{flow}$} &=  \resizebox{0.68\hsize}{!}{$\sum\nolimits{\psi(\textbf{\emph{X}}_1 - \textbf{\emph{w}}(\textbf{\emph{X}}_2))}\odot(1-O_f)/\sum\nolimits{(1-O_f)}$} \\
  &\resizebox{0.7\hsize}{!}{$+\sum\nolimits{\psi(\textbf{\emph{X}}_2 - \textbf{\emph{w}}(\textbf{\emph{X}}_1))}\odot(1-O_b)/\sum\nolimits{(1-O_b)},$}
  \label{eqa:photo_loss}
\end{aligned}
\end{equation}
where \emph{\textbf{w}} is warping operator, $\psi$ is a sparse $L_p$ norm ($p = 0.4$), $O_f$ and $O_b$ are the forward and backward occlusion mask by checking forward-backward consistency, and $\odot$ is a matrix element-wise multiplication. On one hand, the photometric loss $\mathcal{L}^{prior}_{flow}$ facilitates the network to learn the optical flow from the clean domain. On the other hand, the consistency losses $\mathcal{L}^{consis}_{flow}$,  $\mathcal{L}^{align}_{feat}$ benefit the network to transfer the motion from the clean domain to the degraded domain.

% continue
\subsection{Boundary Contrastive Adaptation}
\label{sec:warperror}
% new version
The motion adaptation MCA could offer a preliminary optical flow under adverse weather. However, the motion boundary may be over-smooth and suffers from the perturbation caused by degradation. To further refine the motion boundary, we propose a boundary contrastive adaptation to distinguish the image motion boundary from that of the erroneousness caused by the adverse degradation, such as rain streaks.

In Eq. (\ref{eq:Unconstrained MAP_model}), the ${P}_d(F(D))$ is the prior knowledge about the degradation flow $F(D)$ in a transform domain, which could better facilitate us to decouple the optical flow $F(D)$ from $F(Y)$. Thus, the key is how to properly define the loss about ${P}_d(F(D))$ and design the corresponding network. Since warp error measures the error of motion boundary, we assume the sparsity transform ${P}_d$ as the warp error transform:
\begin{equation}
	\setlength{\abovedisplayskip}{3pt}
	\setlength{\belowdisplayskip}{3pt}
{\textbf{\emph{w}}}(i) = {\textbf{\emph{I}}_1}(i) - {\textbf{\emph{I}}_2}(i + F(i)),
\end{equation}
where $\emph{\textbf{w}}$ is warp error operator, ${\textbf{\emph{I}}_1}$, ${\textbf{\emph{I}}_2}$ are adjacent frames, \emph{F} is optical flow, and \emph{i} is pixel index. Thus the ${P}_d(F(\emph{\textbf{D}}))$ can be expressed as $S(\textbf{\emph{w}}_d)$ where $S(\cdot)$ is a sparsity function, and $\textbf{\emph{w}}_d$ is the warp error about the degradation \textbf{\emph{D}}.

According to the optical flow decomposition model Eq. (\ref{eq:opticalflow_model}), we can further exploit the relationship among warp errors of clean images, degraded images, and degradation as:
\begin{equation}\footnotesize
	\setlength{\abovedisplayskip}{3pt}
	\setlength{\belowdisplayskip}{3pt}
\begin{aligned}
  {\textbf{\emph{w}}_y}(i) - {\textbf{\emph{w}}_x}(i) &= \resizebox{0.62\hsize}{!}{$({\textbf{\emph{Y}}_1}(i) - {\textbf{\emph{X}}_1}(i)) - ({\textbf{\emph{Y}}_2}(i + F(i)) - {\textbf{\emph{X}}_2}(i + F(i)))$}\\
  & = {\emph{\textbf{D}}_1}(i) - {\textbf{\emph{D}}_2}(i + F(i)) = {\textbf{\emph{w}}_d}(i),
\end{aligned}
  \label{eq:warp_error_relationship}
\end{equation}
where $\emph{\textbf{w}}_x$, $\emph{\textbf{w}}_y$ and $\emph{\textbf{w}}_d$ mean the warp errors of each component. Eq. (\ref{eq:warp_error_relationship}) supports our observation that the warp errors between clean domain and degraded domain are inconsistent with $\emph{w}_d$ caused by adverse weather components.
Therefore, instead of directly modeling $\textbf{\emph{w}}_d$, we can indirectly model the warp error relationship between clean domain and degraded domain, namely ($\textbf{\emph{w}}_y - \textbf{\emph{w}}_x$). We take $S(\cdot)$ as exponential function $\exp(\cdot)$, thus ${P}_d(F(\emph{\textbf{D}}))$ can be deduced as:
\begin{equation}
	\setlength{\abovedisplayskip}{3pt}
	\setlength{\belowdisplayskip}{3pt}
{P}_d(F(\emph{\textbf{D}})) = \exp{({\textbf{\emph{w}}}_y -{\textbf{\emph{w}}}_x)} = {\exp({\textbf{\emph{w}}_y}})\backslash{\exp({\textbf{\emph{w}}_x}}).
  \label{eqa:boundary_loss}
\end{equation}
That is to say, ${P}_d(F(\emph{\textbf{D}}))$ is to model the boundary inconsistency between clean and degraded domains. Interestingly, we observe that Eq. (\ref{eqa:boundary_loss}) shares similar formulation with the contrastive learning \cite{chen2020simple}. Both of them are to measure the exclusive relationship between two domains, motivating us to employ the contrastive for boundary adaptation.

In Fig. \ref{Framework}, there are four warp error maps including both the clean and degraded from two different scenes. We design the intra- and inter-scene boundary contrastive adaptation. The former intra-scene adaptation is to learn discriminative motion boundary representation, while the latter inter-scene adaptation is to align boundary manifold variation for different scenes. Both the intra- and inter-scene boundary contrastive adaptations benefit us to better distinguish the clean motion boundary from the degradation motion boundary.

\begin{figure*}
	\setlength{\abovecaptionskip}{8pt}
   \setlength{\belowcaptionskip}{-5pt}
  \centering
  \includegraphics[width=1.0\linewidth]{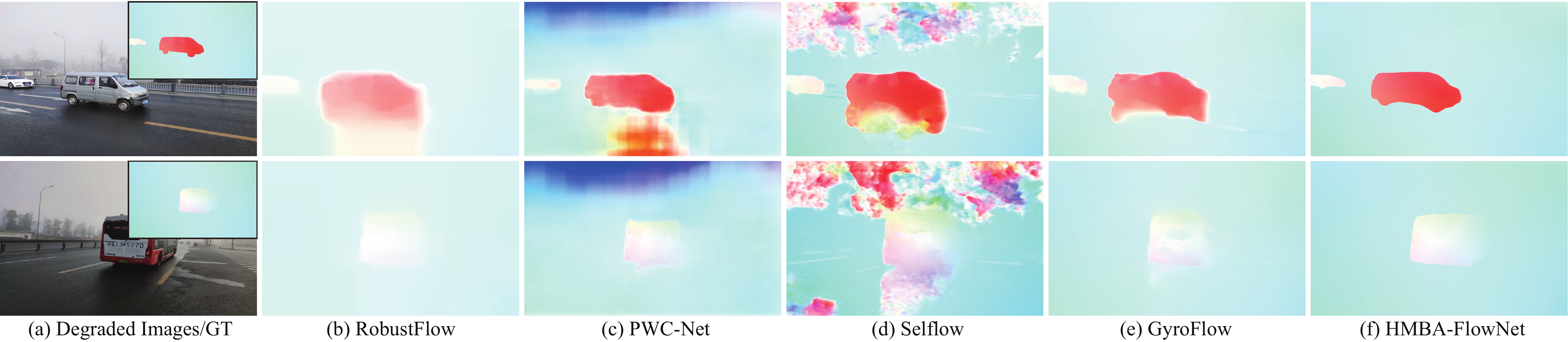}
  \caption{Visual comparison of optical flows on real GOF dataset.}
  \label{Gof_Comparison}
\end{figure*}

\begin{table*}
\footnotesize
\setlength{\belowcaptionskip}{-5pt}
\setlength\tabcolsep{3pt}
\centering
\renewcommand\arraystretch{1.1}
\begin{tabular}{cc|cccccccccc}
\hline
\hline
\multicolumn{2}{c|}{\multirow{2}{*}{Method}}& \multicolumn{1}{c|}{\multirow{2}{*}{RobustFlow}} &\multicolumn{1}{c|}{\multirow{2}{*}{Selflow}}&
\multicolumn{3}{c|}{Selflow} & \multicolumn{1}{c|}{\multirow{2}{*}{ARFlow}} & \multicolumn{3}{c|}{ARFlow} & \multirow{2}{*}{HMBA-Flow}\\

\cline{5-7}\cline{9-11}
\multicolumn{2}{c|}{}& \multicolumn{1}{c|}{} &\multicolumn{1}{c|}{}& \multicolumn{1}{c|}{DDN+} & \multicolumn{1}{c|}{JORDER-E+}& \multicolumn{1}{c|}{RLNet+}& \multicolumn{1}{c|}{}& \multicolumn{1}{c|}{DDN+}& \multicolumn{1}{c|}{JORDER-E+}& \multicolumn{1}{c|}{RLNet+} \\
 \hline
\multicolumn{1}{c|}{\multirow{2}{*}{LR-KITTI}} & EPE & \multicolumn{1}{c|}{22.42}&\multicolumn{1}{c|}{9.07}& \multicolumn{1}{c|}{7.67}& \multicolumn{1}{c|}{6.98}& \multicolumn{1}{c|}{7.13}& \multicolumn{1}{c|}{7.10}& \multicolumn{1}{c|}{7.44}& \multicolumn{1}{c|}{7.65}& \multicolumn{1}{c|}{7.90}& \textbf{5.86} \\
\cline{2-12}
\multicolumn{1}{c|}{}& F1-all & \multicolumn{1}{c|}{80.13\%} & \multicolumn{1}{c|}{44.87\%}& \multicolumn{1}{c|}{44.42\%} & \multicolumn{1}{c|}{42.08\%} & \multicolumn{1}{c|}{42.14\%} & \multicolumn{1}{c|}{43.47\%} & \multicolumn{1}{c|}{44.68\%} & \multicolumn{1}{c|}{43.86\%} & \multicolumn{1}{c|}{44.95\%} & \textbf{33.61\%} \\
\hline
\multicolumn{1}{c|}{\multirow{2}{*}{HR-KITTI}} & EPE & \multicolumn{1}{c|}{23.40} & \multicolumn{1}{c|}{10.78}& \multicolumn{1}{c|}{7.82}& \multicolumn{1}{c|}{7.11}& \multicolumn{1}{c|}{7.04}& \multicolumn{1}{c|}{8.89}& \multicolumn{1}{c|}{7.74}& \multicolumn{1}{c|}{6.95}& \multicolumn{1}{c|}{7.77}& \textbf{6.47}\\
\cline{2-12}
\multicolumn{1}{c|}{}& F1-all & \multicolumn{1}{c|}{81.12\%} & \multicolumn{1}{c|}{52.18\%}& \multicolumn{1}{c|}{45.43\%} & \multicolumn{1}{c|}{41.78\%} & \multicolumn{1}{c|}{41.35\%} & \multicolumn{1}{c|}{48.98\%} & \multicolumn{1}{c|}{45.05\%} & \multicolumn{1}{c|}{43.55\%} & \multicolumn{1}{c|}{45.89\%} & \textbf{38.19\%}  \\
\hline
\hline
\end{tabular}
\caption{Quantitative results on synthetic Heavy (HR-KITTI) and Light Rain-KITTI2015 (LR-KITTI) datasets.}
 \label{tab:quantitative_result}
\end{table*}

In contrastive adaptation, we propose an edge-aware sampling strategy by extracting the salient edges \cite{dollar2013structured} [green patches in Fig. \ref{Framework}] from the clean domain as the positive samples, in order to get rid of the most meaningless region without zero values. As for the negative samples, we propose a local entropy-aware sampling strategy by extracting the salient edges [pink patches in Fig. \ref{Framework}] from the degraded domain. We sort the local entropy (degree of degradation) in descending order and choose the top \emph{N} patches. Note that, the negative patches would be sampled with different locations from the positives, so as to exclude the clean image boundary. The advantage of the proposed sampling strategy over the conventional random sampling strategy is analyzed in the discussion.

%As for the encoder, our goal is to extract discriminative features of warp error map for determining the alignment of two adjacent frames that are affected by degradation or not. The generator $G$ is to learn how to generate degradation, which naturally knows the property of degradation. Thus, we choose the encoder of \emph{G} as the feature encoder $E_G$.

Therefore, we denoted \resizebox{0.58\hsize}{!}{$f_{P_{j}^x} = E_G(w_{x_j})$, $f_{P_{j}^{y}} = E_G(w_{y_j})$} and \resizebox{0.25\hsize}{!}{$f_{N_{i}} = E_G(w_{y_i})$} for the output positive and negative features. In the boundary adaptation step, the intra-scene contrastive adaptation loss is as follows:
\begin{equation}
	\setlength\abovedisplayskip{1.5pt}
  \setlength\belowdisplayskip{1.5pt}
\begin{aligned}
  \resizebox{0.88\hsize}{!}{$\mathcal{L}^{contra}_{errIntra} = \sum\nolimits_{j=1}^N \frac{exp(f_{P_{j}^x}\cdot f_{P_{j}^y} / \tau)}{exp(f_{P_{j}^x}\cdot f_{P_{j}^y} / \tau)+\sum\nolimits_{i=1}^N{exp(f_{N_{i}}\cdot f_{P_{j}^x}/\tau)}},$}
  \label{eqa:contra_loss}
\end{aligned}
\end{equation}
The intra-scene contrastive loss constrains the clean warp error patches $\textbf{\emph{w}}_{x_j}$ at location $\emph{j}$ to be positive with the corresponding degraded patches $\textbf{\emph{w}}_{y_j}$ in comparison to other warp error patches $\textbf{\emph{w}}_{y_i}$, so as to force motion boundary of the degraded domain close to that of the clean domain in the same scene and get rid of degradation. We further enforce the inter-scene contrastive adaptation loss to align the distribution of motion boundary in the clean domain of different scenes:
\begin{equation}
  \setlength\abovedisplayskip{3pt}
  \setlength\belowdisplayskip{3pt}
\begin{aligned}
  \resizebox{0.88\hsize}{!}{$\mathcal{L}^{contra}_{errInter} = \frac{1}{N} \sum\nolimits_{k=1}^N \sum\nolimits_{j=1}^N\frac{exp(f_{P_{j}^x}\cdot f_{P_{k}^x} / \tau)}{exp(f_{P_{j}^x}\cdot f_{P_{k}^x} / \tau)+\sum\nolimits_{i=1}^{2N}{exp(f_{N_{i}}\cdot f_{P_{j}^x}/\tau)}},$}
  \label{eqa:contra_loss}
\end{aligned}
\end{equation}
where \emph{N} is the positive/negative sample numbers,  $\tau$ denotes the scale parameter.
The total loss of our framework strictly mimics the optimization function Eq. (\ref{eq:Unconstrained MAP_model}) as follows:
\begin{equation}
	\setlength\abovedisplayskip{2pt}
  \setlength\belowdisplayskip{2pt}
	\begin{aligned}
  \mathcal{L} = \alpha\mathcal{L}^{tran}_{img} + \beta\mathcal{L}^{prior}_{img} +  \gamma_{1}\mathcal{L}^{align}_{feat} + \gamma_{2}\mathcal{L}^{consis}_{flow} \\
	+ \delta\mathcal{L}^{prior}_{flow} + \rho_{1}\mathcal{L}^{contra}_{errIntra} + \rho_{2}\mathcal{L}^{contra}_{errInter},
  \label{total_loss}
\end{aligned}
\end{equation}

\subsection{Implementation Detail}
We empirically set the parameters $\{\alpha,\beta,\gamma_{1},\gamma_{2},\delta,\rho_{1},\rho_{2}\}$ = $ \{1,1,1,1,1,0.1,0.1\}$. Our method is implemented on the Tensorflow platform with two NVIDIA RTX 2080Ti GPUs within three days. Note that, the proposed network contains image translation generators and discriminators, two unshared optical flow encoders and one shared decoder, and one shared contrastive encoder. We first train the image translation generators and discriminators via $\mathcal{L}^{tran}_{img}$ and $\mathcal{L}^{prior}_{img}$.
Then we update the optical flow encoders and decoder via $\mathcal{L}^{align}_{feat}$, $\mathcal{L}^{consis}_{flow}$ and $\mathcal{L}^{prior}_{flow}$ with 200k iterators and 0.0002 learning rate, where we initialize the optical flow network of clean domain with clean KITTI2015 dataset \cite{menze2015object} via $\mathcal{L}^{prior}_{flow}$ only.
After that, we learn the whole network with the contrastive losses $\mathcal{L}^{contra}_{errIntra}$ and $\mathcal{L}^{contra}_{errInter}$.
At the test stage, the final model only needs the optical flow encoder $E_y$ and decoder $D$ which takes real degraded images as input and outputs corresponding optical flow.

\begin{table}
\footnotesize
\setlength{\belowcaptionskip}{-5pt}
	\setlength\tabcolsep{2pt}
\centering
  \renewcommand\arraystretch{1.2}
\begin{tabular}{cc|ccccc}
	\hline
	\hline
	\multicolumn{2}{c|}{Method}& \multicolumn{1}{c|}{RobustFlow} &\multicolumn{1}{c|}{PWC-Net}  & \multicolumn{1}{c|}{Selflow}  & \multicolumn{1}{c|}{GyroFlow} & \multicolumn{1}{c}{Ours}  \\
	\hline
	\multicolumn{1}{c|}{\multirow{2}{*}{Rain}} & EPE & \multicolumn{1}{c|}{7.10}&\multicolumn{1}{c|}{2.85}& \multicolumn{1}{c|}{1.78}& \multicolumn{1}{c|}{1.07}& \textbf{0.91} \\
	\cline{2-7}
	\multicolumn{1}{c|}{}& F1-all & \multicolumn{1}{c|}{53.35\%} & \multicolumn{1}{c|}{21.39\%}& \multicolumn{1}{c|}{16.21\%} & \multicolumn{1}{c|}{9.94\%} & \textbf{8.07\%} \\
	\hline

	\multicolumn{1}{c|}{\multirow{2}{*}{Fog}} & EPE & \multicolumn{1}{c|}{12.25}&\multicolumn{1}{c|}{5.99}& \multicolumn{1}{c|}{1.97}& \multicolumn{1}{c|}{0.95}& \textbf{0.84} \\
	\cline{2-7}
	\multicolumn{1}{c|}{}& F1-all & \multicolumn{1}{c|}{80.93\%} & \multicolumn{1}{c|}{41.02\%}& \multicolumn{1}{c|}{18.35\%} & \multicolumn{1}{c|}{9.13\%} & \textbf{7.46\%} \\
	\hline
	\hline
 \end{tabular}
 \caption{Quantitative results on real GOF dataset.}
 \label{tab:gof_comparison}
\end{table}

\section{Experiments}
\subsection{Experiment Setup}
\textbf{Datasets.} We validate the performance on one synthetic and two real degraded datasets. We simulate the synthetic rain with different densities of rain (\emph{e.g.}, heavy rain and light rain) on KITTI2015 by the software \emph{Adobe After Effects}. GOF \cite{li2021gyroflow} is a real dataset with 1000 images for training and 120 images for testing. Another real adverse weather dataset is collected from the \emph{Youtube}. We select 1200 adverse weather images including \emph{e.g.}, rain, fog, and snow for training and 200 images for testing.
We choose EPE \cite{dosovitskiy2015flownet} and F1-all \cite{menze2015object} for the quantitative evaluation.

\begin{figure*}
	\setlength{\abovecaptionskip}{8pt}
   \setlength{\belowcaptionskip}{-5pt}
  \includegraphics[width=1.0\linewidth]{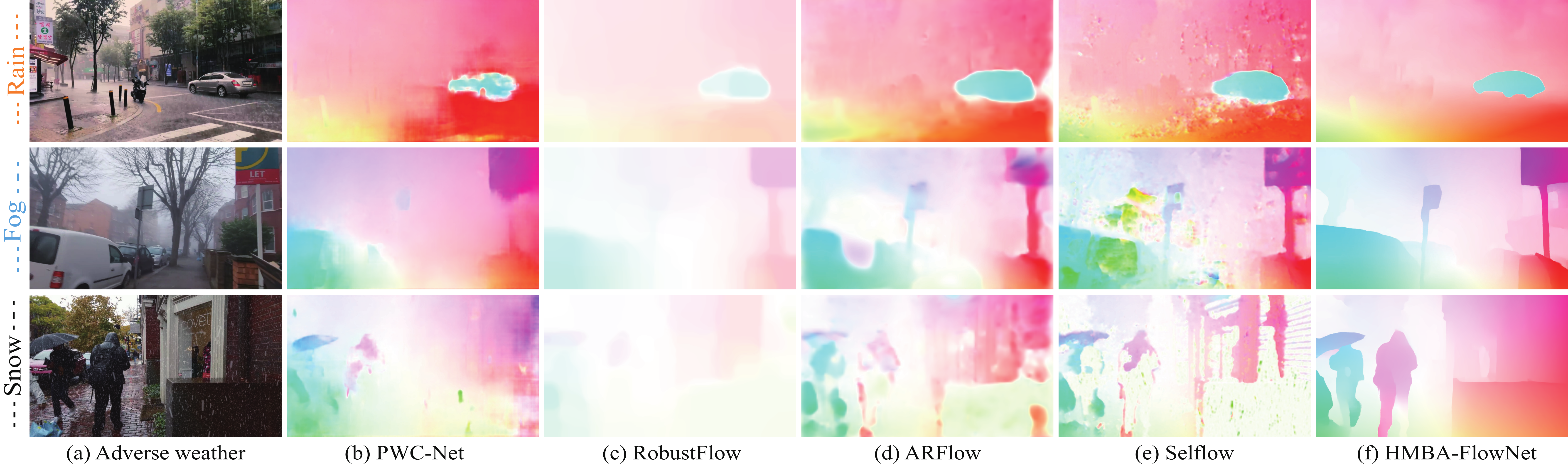}
  % \vspace*{-2mm}
  \caption{Comparison of optical flows on real adverse weather dataset. HMBA-FlowNet can estimate motion-smooth and boundary-sharpen optical flow under adverse weather. Moreover, it can generalize well for real various adverse weather.}
  \label{fig:real_data_experiment}
  % \vspace*{-2.5mm}
\end{figure*}

\noindent
\textbf{Comparison Methods.} Two competing methods weakly-supervised GyroFlow \cite{li2021gyroflow} and optimization-based RobustFlow \cite{li2018robust} are chosen which are designed for rainy scene optical flow. Moreover, we select several state-of-the-art supervised (PWC-Net \cite{sun2018pwc}) and unsupervised (Selflow \cite{liu2019selflow} and ARFlow \cite{liu2020learning}) clean image-based optical flow approaches.
For fair comparison, the supervised methods are only trained on the synthetic dataset with optical flow labels. The unsupervised methods only have access to the degraded images for training.
As for the synthetic dataset, we design two different training strategies for competing methods. The first one is that we directly train the comparison methods on rainy images. The second one is to perform the deraining first via derain approaches (\emph{e.g.}, DDN \cite{fu2017removing}, JORDER-E \cite{8627954} and RLNet \cite{9577602}), and then we train the comparison methods on the deraining results (named as {DDN+/JORDER-E+/RLNet+}).

\begin{figure}[t]
	\setlength{\abovecaptionskip}{8pt}
   \setlength{\belowcaptionskip}{-5pt}
  \includegraphics[width=0.99\linewidth]{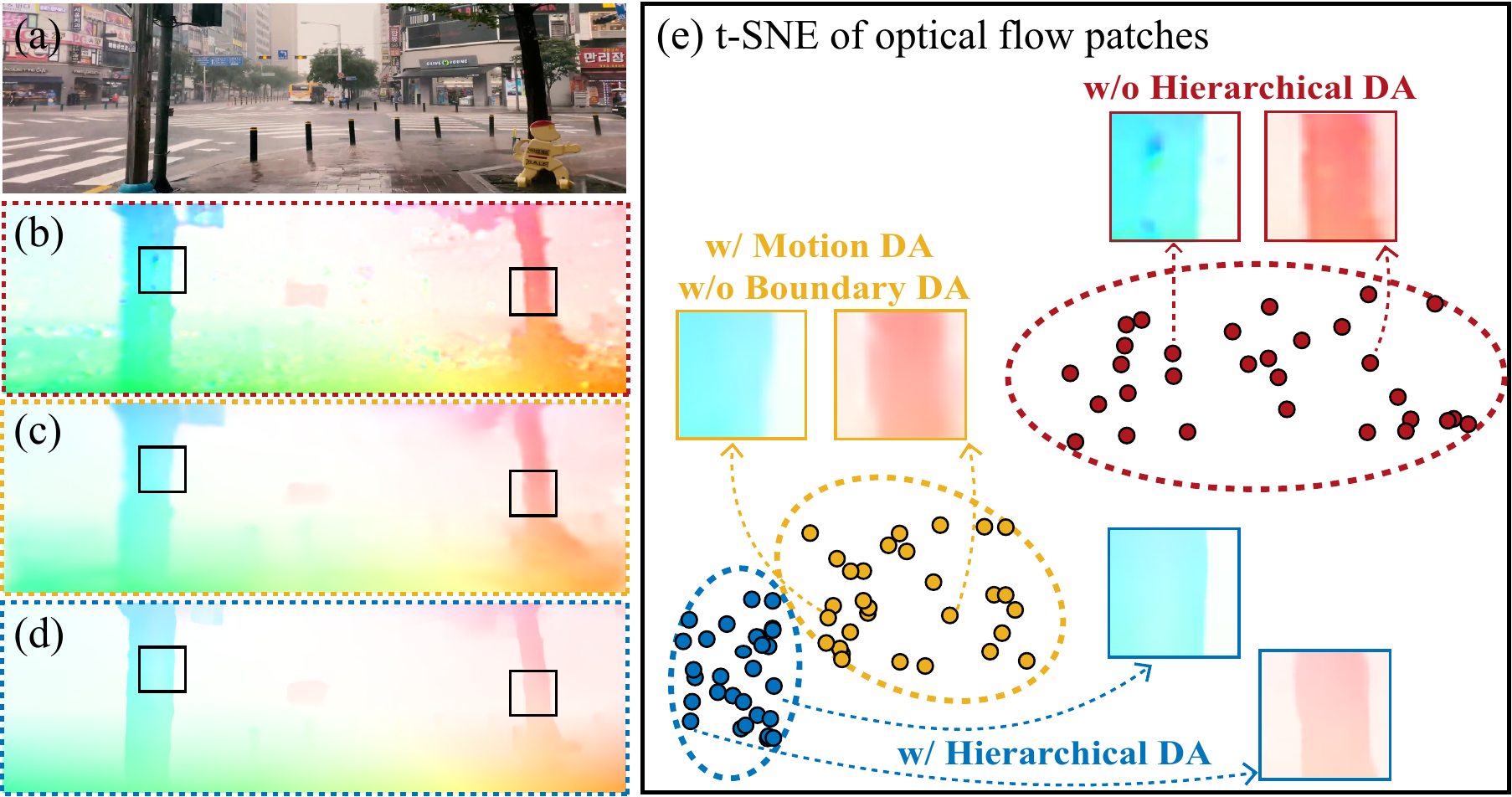}
  \caption{Effectiveness of hierarchical adaptation architecture. (a) Rainy image. (b)-(d) Optical flows without DA, with MCA only, with hierarchical MCA and BCA, respectively. (e) The t-SNE visualization of each adaptation strategy.}
  \label{fig:ablation_cl}
\end{figure}

\subsection{Experiments on Synthetic Images}
In Table \ref{tab:quantitative_result}, we show the quantitative comparison on the synthetic heavy and light Rain-KITTI2015 dataset. We have two key observations. First, the proposed HMBA-FlowNet is significantly better than that of the unsupervised counterparts under different rainy conditions. Second, the pre-processing procedure deraining may not be always positive to the final optical flow. As for the light rain condition, the deraining result may be even harmful because the image deraining would unexpectedly cause damage to image details and outweigh its removal benefits. On the contrary, the proposed method could well handle both the light and heavy rain conditions, since the motion-consistency and boundary-inconsistency knowledge would greatly relieve the need for the clean image.

\subsection{Experiments on Real Images}
In Fig. \ref{Gof_Comparison} and \ref{fig:real_data_experiment}, we show the visual comparison results on real GOF and collected datasets, respectively. The supervised PWC-Net contains obvious erroneousness, due to the domain gap between synthetic training and real test images. The unsupervised methods Selflow and ARFlow cannot work well, in which there are artifacts because degradation has violated the basic assumptions of optical flow. The hardware-assisted GyroFlow can predict the accurate motion of the background, but the boundary of the foreground moving object is not clear in Fig. \ref{Gof_Comparison}(e), since it heavily relies on the ego-motion of the camera captured by the gyroscope data yet less effective for the independent foreground object motion.
The proposed HMBA-FlowNet can remove erroneous outliers and obtain satisfactory results under real adverse weather in Fig. \ref{Gof_Comparison} and \ref{fig:real_data_experiment}(f). In Table \ref{tab:gof_comparison}, the quantitative results on GOF dataset further verify the superiority of the proposed method.

\begin{table}[t]
\footnotesize
\setlength{\belowcaptionskip}{-7pt}
\setlength\tabcolsep{3pt}
\centering
   \renewcommand\arraystretch{1.1}
\begin{tabular}{cccc|cc}
               \hline
				$\mathcal{L}^{consis}_{flow}$&$\mathcal{L}^{align}_{feat}$&$\mathcal{L}^{contra}_{errIntra}$ &$\mathcal{L}^{contra}_{errInter}$ & LR-KITTI & HR-KITTI \\
							 \hline
	         $\times$  &$\times$&$\times$& $\times$& 9.14 & 10.61  \\

			 $\surd$& $\times$&$\times$&  $\times$& 6.71 & 6.89 \\

			$\surd$ 	&$\surd$ &$\times$&$\times$& 6.65 & 6.89 \\

			$\surd$ &$\surd$ &$\surd$ & $\times$& 6.52 & 6.67 \\
			$\surd$ 	&$\surd$ &$\times$&$\surd$  & 6.01 & 6.54 \\
			$\surd$ 	&$\surd$ &$\surd$ &$\surd$ & \textbf{5.86} &\textbf{6.47} \\
         \hline
 \end{tabular}
 \caption{Ablation study on motion-boundary adaptation loss.}
 \label{tab:ablation_cl}
\end{table}

\subsection{Ablation Study}
\textbf{Effectiveness of Hierarchical Adaptation Architecture.} To illustrate the effectiveness of the hierarchical adaptation architecture, in Fig. \ref{fig:ablation_cl}, we show the optical flow estimation of different architectures and visualize their low-dimensional distributions via t-SNE. In Fig. \ref{fig:ablation_cl}(b), we can observe that there are obvious artifacts in the optical flow without hierarchical adaptation. With motion adaptation only [Fig. \ref{fig:ablation_cl}(c)], most of the outliers caused by degradation are removed. With the additional boundary adaptation [Fig. \ref{fig:ablation_cl}(d)], the flow boundary is significantly clearer.
We can conclude that the hierarchical domain adaptation could improve adverse weather optical flow. Moreover, in Fig. \ref{fig:ablation_cl}(e), we visualize their corresponding t-SNE distributions. The red, yellow, and blue denote the without hierarchical adaptation, with motion adaption yet without boundary adaptation, and with hierarchical motion and boundary adaptation. The red circle is scattered, the yellow circle is gradually focused, and the blue circle is most concentrated, illustrating that hierarchical adaptation could greatly improve optical flow discriminative representation.

\noindent
\textbf{Effectiveness of Motion-Boundary Adaptation Losses.} In Table \ref{tab:ablation_cl}, we show how motion-boundary adaptation losses contribute to the final result. \resizebox{0.10\hsize}{!}{$\mathcal{L}^{align}_{feat}$} and \resizebox{0.12\hsize}{!}{$\mathcal{L}^{consis}_{flow}$} aim to transfer optical flow knowledge from the clean domain to the degraded domain via the consistency loss.
The goal of  $\mathcal{L}^{contra}_{errIntra}$ and $\mathcal{L}^{contra}_{errInter}$ is to distinguish the motion boundary of image structure from that of the degradation errors, and also align the motion boundary distribution of different scenes. We can observe that motion consistency losses make a major contribution to the optical flow results and boundary contrastive losses further improve the final results.

\begin{table}[t]
\footnotesize
\setlength\tabcolsep{3pt}
\setlength{\belowcaptionskip}{-7pt}
\centering
  \renewcommand\arraystretch{1.15}
\begin{tabular}{c|c|c|c|c}
               \hline
							 \hline
							  Domain & \multicolumn{1}{c|}{\multirow{1}{*}{Clean}} &
							 \multicolumn{1}{c|}{\multirow{1}{*}{Degraded}}&
							  \multicolumn{1}{c|}{\multirow{1}{*}{LR-KITTI}} & \multicolumn{1}{c}{\multirow{1}{*}{HR-KITTI}} \\

							 \hline
							 \multicolumn{1}{c|}{\multirow{4}{*}{\makecell{Sampling \\ strategy}}}& Random & Random & 6.59 & 5.74 \\
							 \cline{2-5}

							 &Random & Entropy-aware & 6.43 & 6.72 \\
							 \cline{2-5}
							 &Edge-aware & Random & 5.98 & 6.52 \\
							 \cline{2-5}
							 &Edge-aware & Entropy-aware & \textbf{5.86} & \textbf{6.47} \\

         \hline
				 \hline
				 & \multicolumn{2}{c|}{\multirow{1}{*}{Optical flow encoder}} &
				 \multicolumn{1}{c|}{\multirow{1}{*}{LR-KITTI}} & \multicolumn{1}{c}{\multirow{1}{*}{HR-KITTI}} \\

				 \hline
				 \multicolumn{1}{c|}{\multirow{2}{*}{\makecell{Weight \\ strategy}}}& \multicolumn{2}{c|}{\multirow{1}{*}{Shared}} & 7.32 & 7.60 \\
				 \cline{2-5}
				 & \multicolumn{2}{c|}{\multirow{1}{*}{Unshared (ours)}} & \textbf{5.86} & \textbf{6.47} \\

				 \hline
				 \hline
				 & \multicolumn{2}{c|}{\multirow{1}{*}{Image translation network}} &
				 \multicolumn{1}{c|}{\multirow{1}{*}{LR-KITTI}} & \multicolumn{1}{c}{\multirow{1}{*}{HR-KITTI}} \\

				 \hline
				 \multicolumn{1}{c|}{\multirow{2}{*}{\makecell{Backbone \\ strategy}}}& \multicolumn{2}{c|}{\multirow{1}{*}{PatchGAN}} & 6.67 & 7.03 \\
				 \cline{2-5}
				 & \multicolumn{2}{c|}{\multirow{1}{*}{CycleGAN (ours)}} & \textbf{5.86} & \textbf{6.47} \\

				 \hline
				 \hline
 \end{tabular}
 \caption{Discussion on strategies of different modules.}
 \label{tab:discussion}
 % \vspace{-5mm}
\end{table}

\subsection{Discussion}
\textbf{Contrastive Sampling Strategies.} The positive and negative sampling strategy is of great importance to the quality of the discriminative feature in contrastive learning. In Table \ref{tab:discussion}, we show the advantage of the proposed edge-aware and entropy-aware sampling strategies over the conventional random sampling. Compared with random sampling, the edge-aware sampling strategy in the clean domain could greatly improve the optical flow. The main reason is most random sample patches in the warp error map are the meaningless region with zero values. On the contrary, the sharp edge-aware sampling would guarantee us abundant discriminative information so as to better reflect the alignment of the optical flow boundary. Similarly, the entropy-aware sampling could also select the informative highly-informative boundary of warp error caused by degradations, facilitating to differ the flow boundary of the image structure from that of degradations.

\noindent
\textbf{Shared v.s. Unshared Weight.} To study the influence of flow encoders, in Table \ref{tab:discussion}, we design the shared- and unshared-weight for the optical flow encoders. We can observe that the performance of the unshared-weight strategy is better than that of the shared-weight strategy. Imagine that if the two flow encoders are exactly shared, the learned features are hard to be aligned into the common space due to the different inputs of the two domains.
The optical flow of clean domain is learned via the photometric loss, and transferred to the degraded domain via the consistency loss. Thus, it is natural to enforce different learning weights for the two encoders.

\begin{figure}[t]
	\setlength{\abovecaptionskip}{8pt}
   \setlength{\belowcaptionskip}{-5pt}
\centering
  \includegraphics[width=0.99\linewidth]{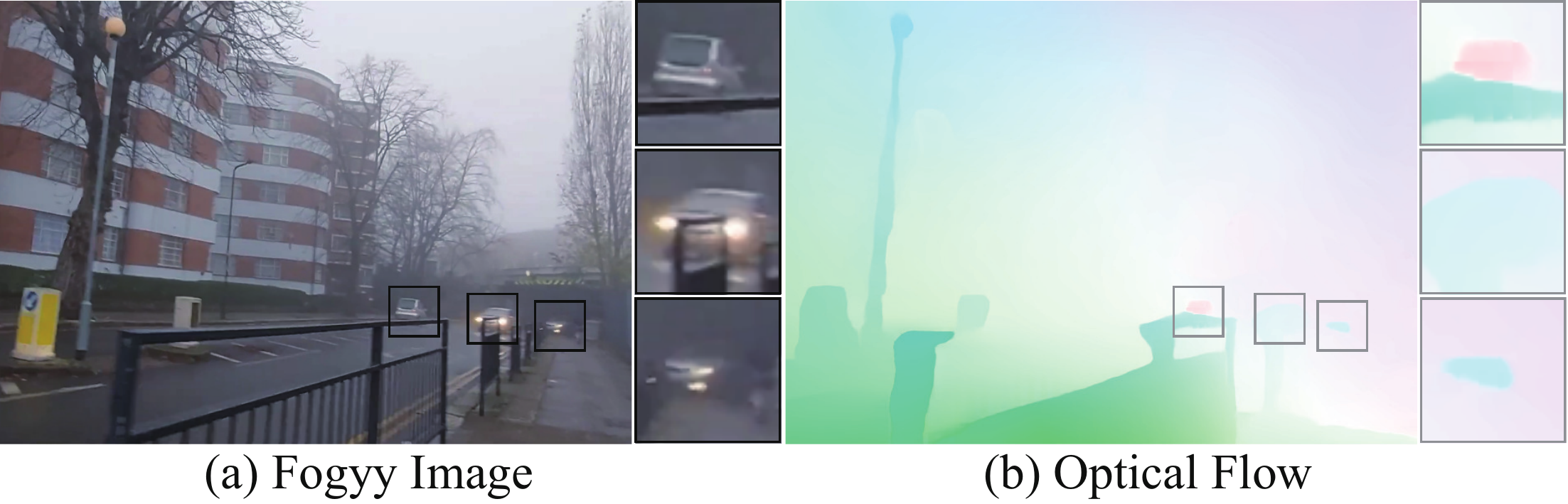}
  \caption{The proposed method still faces great challenging under dense foggy, especially for objects in distant regions.}
  \label{fig:limitations}
\end{figure}

\noindent
\textbf{Choice of the Image Translation Network.} The key to the proposed method is to build the relationship between the clean and degraded domains. Thus, the image translation between the two domains also plays an important role. We have tested the classical PatchGAN \cite{isola2017image} and also the CycleGAN \cite{zhu2017unpaired} as the baseline, as shown in Table \ref{tab:discussion}. It is shown that the CycleGAN is slightly better than the PatchGAN. Thus, we choose CycleGAN as our image translation network. Note that, the proposed method utilizes the self-supervised motion consistency and boundary inconsistency relationship between both domains, which significantly relieves the requirement of absolute clean images.

\noindent
\textbf{Limitation.} The HMBA-FlowNet may fail for distant objects under dense fog. In Fig. \ref{fig:limitations}, although our method can well estimate the optical flow of the nearby object, for the distant object our method may suffer from challenges. In this work, we simply assume the adverse artifacts obey a linear additive model via Eq. (\ref{eq:linear_model}). However, fog is a non-uniform degradation highly related to scene depth, in which degraded optical flow at the nearby distance can be approximated, but flow in the distant area does not satisfy the degradation model. In the future, we will incorporate depth into our degradation model.

\section{Conclusion}
In this work, we propose the first unsupervised hierarchical motion-boundary domain adaptation framework for adverse weather optical flow. We mathematically reveal that optical flow consistency and warp error inconsistency between clean and degraded domains can be well utilized for optical flow estimation. Thus, we transfer knowledge from clean domain to degraded domain by learning a motion-invariance common space with the consistency constraint. Moreover, we refine the motion boundary via boundary contrastive adaptation to effectively distinguish the motion boundary from the warp error caused by degradation. The hierarchy motion-boundary adaptation benefits us to handle the various adverse weather optical flow. We have conducted extensive experiments to verify the superiority of the proposed HMBA-FlowNet.

\section{Acknowledgments}
This work was supported in part by the National Natural Science Foundation of China under Grant 61971460, in part by JCJQ Program under Grant 2021-JCJQ-JJ-0060, and in part by the National Natural Science Foundation of China under Grant 62101294.

\bibliography{aaai23}

\end{document}